%% file: main.tex
\documentclass[runningheads]{llncs}
\usepackage[T1]{fontenc}
\usepackage{graphicx}
\usepackage{makecell}
\usepackage{bigstrut}
\usepackage{amsmath,amssymb,amsfonts}
\usepackage{xspace}
\usepackage{booktabs}
\usepackage{comment}
\usepackage{threeparttable}
\usepackage{subcaption}
\usepackage{gnuplottex}
\usepackage{multicol}
\usepackage{soul}
\usepackage{enumitem}
\usepackage{xcolor}
\usepackage{wrapfig}
\usepackage{caption}
\usepackage{hyperref}
\usepackage[misc]{ifsym}
\usepackage{color}

\input{commands}

\graphicspath{{figures/}}

\begin{document}
\title{CoMAC: Conversational Agent for Multi-Source Auxiliary Context with Sparse and Symmetric Latent Interactions}
\titlerunning{CoMAC: Conversational Agent for Multi-Source Auxiliary Context}
\author{
\Letter~Junfeng Liu\inst{1,2} \and
Christopher T. Symons\inst{2} \and
Ranga Raju Vatsavai\inst{1,2}
}
\authorrunning{J. Liu et al.}
% \author{Anonymous}
% \authorrunning{Anonymous}

%
\institute{
North Carolina State University, Raleigh, NC 27606, USA \and
Lirio, Inc., Knoxville, TN 37923, USA \\
\email{jliu85@ncsu.edu},
\email{csymons@lirio.com},
\email{rrvatsav@ncsu.edu}
}
% \institute{Anonymous Institute}

%
\maketitle              % typeset the header of the contribution

\input{sections/abstract}
\input{sections/intro}
\input{sections/related_work}

\input{sections/method}
\input{sections/experiments}

\input{sections/results}
\input{sections/ablation}

\input{sections/conclusion}

\begin{credits}
\subsubsection{\ackname} \newcontent{We would like to thank Joongyeub Yeo (Lirio) and Seoyeong Park (NCSU) for their valuable suggestions on an early version of this paper.}

\textbf{Disclosure:} This paper has been accepted to The 29th Pacific-Asia Conference on Knowledge Discovery and Data Mining (PAKDD2025) and will be published in the conference proceedings by Springer. DOI will provided once available. 
\end{credits}
\bibliographystyle{splncs04}
\bibliography{main}

\end{document}

%% file: commands.tex
\newcommand{\etal}{\textit{et al.}\xspace}

\newcommand{\pkfocus}{\texttt{PK-FoCus}\xspace}
\newcommand{\pkncli}{\texttt{PK-NCLI}\xspace}
\newcommand{\pkfocusO}{\texttt{PK-FoCus}$^0$\xspace}
\newcommand{\pkncliO}{\texttt{PK-NCLI}$^0$\xspace}
\newcommand{\pksncli}{\mbox{\ourmethod-$\sncolbertsim$}\xspace}
\newcommand{\ourmethod}{\texttt{CoMAC}\xspace}

\newcommand{\colbert}{\texttt{ColBERT}\xspace}

\newcommand{\colbertsim}{\mathcal{S}\textsubscript{\texttt{C}}\xspace}
\newcommand{\ncolbertsim}{\mathcal{S}\textsubscript{\texttt{N}}\xspace}
\newcommand{\sncolbertsim}{\mathcal{S}\textsubscript{\texttt{SN}}\xspace}
\newcommand{\ssncolbertsim}{\mathcal{S}\textsubscript{\texttt{SSN}}\xspace}

\newcommand{\sncr}{\mathcal{P}_{sr}}
\newcommand{\answer}{\mathcal{A}}
\newcommand{\answercomac}{\mathcal{A}_{C}}
\newcommand{\answerfocus}{\mathcal{A}_{F}}

\newcommand{\real}{\mathbb{R}\xspace}

\newcommand{\softmax}{\texttt{softmax}\xspace}
\newcommand{\argmax}{\texttt{argmax}\xspace}

\newcommand{\tokenhl}[1]{\textbf{\ul{#1}}\xspace}

\newcommand{\newcontent}[1]{#1\xspace}
\newcommand{\removedcontent}[1]{} 

%% file: sections/abstract.tex
\begin{abstract}

Recent advancements in AI-driven conversational agents have exhibited immense potential of AI applications. Effective response generation is crucial to the success of these agents. While extensive research has focused on leveraging multiple auxiliary data sources (e.g., knowledge bases and personas) to enhance response generation, existing methods often struggle to efficiently extract relevant information from these sources.
There are still clear limitations in the ability to combine versatile conversational capabilities with adherence to known facts and adaptation to large variations in user preferences and belief systems, which continues to hinder the wide adoption of conversational AI tools. 
This paper introduces a novel method, 
\ul{Co}nversational Agent for \ul{M}ulti-Source \ul{A}uxiliary \ul{C}ontext with Sparse and Symmetric Latent Interactions (\ourmethod),
for conversation generation, which employs specialized encoding streams and post-fusion grounding networks for multiple data sources to identify relevant persona and knowledge information for the conversation. \ourmethod also leverages a novel text similarity metric that allows bi-directional information sharing among multiple sources and focuses on a selective subset of meaningful words. 
Our experiments show that \ourmethod improves the relevant persona and knowledge prediction accuracies and response generation quality significantly over two state-of-the-art methods.
% \ourmethod surpasses two state-of-the-art methods in relevant persona and knowledge prediction accuracies by 45.74\% and 6.76\%, respectively. In addition, it enhances response generation quality by 5.26\%, 5.54\%, 7.84\%, and 11.64\% in terms of F1, ROUGE-L, BLEU, and PPL scores.
% This paper further provides a detailed analysis of how sparsity and symmetry impact agent performance and conducts a comprehensive ablation study to evaluate the method's sensitivity to various hyper-parameters.

% \keywords{Conversational agent \and Natural Language Processing}

\end{abstract}

%% file: sections/intro.tex
\section{Introduction}

\newcontent{
Recent advancements in deep learning and large language models (LLMs) are facilitating widespread application of Conversational AI. Many real-world applications demand human-level performance and additional enhancements tailored to specific use cases.
%
% auxiliary data (persona, knowledge) is important to conversational models
Achieving these objectives typically requires carefully curated data, including auxiliary data that provides implicit or explicit context for models to generate better responses~\cite{dinan2018wizard,zhang2018personalizing,mostafazadeh2017image}.
Many applications require conversational agents to provide answers that adhere to known facts. 
 Therefore, it is crucial to effectively utilize auxiliary knowledge bases and extract relevant information when generating responses. 
There is also growing demand for conversational systems that can enhance the user experience via personalization, improving engagement and aiding users in achieving personal goals. This requires models to better understand auxiliary user information that captures meaningful context, such as demographics, preferences, and beliefs.}
For instance, to support different patients from different backgrounds and in different states of health in attaining particular behavioral objectives tied to improved health, an agent must accurately reflect knowledge and facts about specific health conditions, and employ customized persuasive strategies aligned with each individual patient's beliefs and medical history.

\begin{comment}
    Complicating the situation is the unfortunate reality that despite offering valuable insights to enhance a conversational agent's understanding of context, auxiliary information can also introduce noise or irrelevant data that can hinder comprehension. Effectively and efficiently extracting relevant information and filtering out irrelevant noise for response generation remains a challenge. For this reason, addressing these issues remains a vital area of research in conversational AI. 
\end{comment}

Despite the potential to offer valuable insights that can enhance a conversational agent's understanding of context, auxiliary data also has the potential to introduce noise that can hinder comprehension. It still remains a challenge to effectively and efficiently extracting relevant information while filtering out irrelevant data for response generation. For this reason, addressing these issues remains a vital area of research in conversational AI.

To address these challenges, we present a novel conversational method, \ourmethod, that can jointly leverage different types of auxiliary data to enhance the quality of generated responses. 
Specifically, \ourmethod encodes the conversational history, as well as each form of auxiliary data (persona/knowledge) in individual, specialized encoding streams, and adopts separate post-fusion persona and knowledge grounding networks (denoted PG and KG) that effectively utilizes each stream to identify relevant information. Meanwhile, a novel text similarity metric, $\ssncolbertsim$, is developed to exploit low-level word-to-word similarities among multiple sources of auxiliary contexts by introducing normalization, symmetry, and sparsity. 

\ourmethod can be easily extended to other applications with additional auxiliary data sources due to the flexibility of the post-fusion framework. 
We evaluate \ourmethod against two state-of-the-art (SOTA) methods, \pkfocus~\cite{jang2022call} and \pkncli~\cite{liu2023context}, and adopt a new training setup that addresses the inherent data imbalance issues overlooked by previous work. 
Our experiments demonstrated that \ourmethod significantly outperforms the best SOTA \pkncli, in terms of both PG/KG grounding accuracies by 45.74\% and 6.76\%, respectively. In addition, it enhances response generation quality by 5.26\%, 5.54\%, 7.84\%, and 11.64\% in F1, ROUGE-L, BLEU, and PPL scores.

%% file: sections/related_work.tex
\section{Related Work}
\label{sec:related}
\subsection{Neural-Based Conversational Models}
\label{sec:related:neural}

Deep neural network (DNN) based conversational systems have recently become prevalent due to superior scalability and applicability~\cite{gao2018neural} compared to traditional systems. Neural networks can extract signals directly from end-to-end training driven by data without human intervention. With the ever-increasing learning capabilities of LLMs, especially pre-trained Transformer-based LLMs~\cite{vaswani2017attention}, DNN-based agents are being continually developed for new applications. \textit{Retrieval} methods~\cite{humeau2019poly,liu2023pcpe} and \textit{Generative} methods~\cite{zhang2019dialogpt,li2016persona} are two common categories of DNN-based conversational methods. Many document retrieval solutions~\cite{khattab2020colbert} can also be easily adapted to solve conversational challenges.

\subsection{\newcontent{Auxiliary Data-Enhanced Models} \removedcontent{(combines previous two)}}
\label{sec:related:auxiliary}

%\newcontent{Various research studies have focused on better utilizing external knowledge to enhance conversational agents.}
%
Knowledge-based models~\cite{dinan2018wizard,ghazvininejad2018knowledge} are focused on better utilizing auxiliary facts to enhance the accuracy of responses. Knowledge and facts are crucial for many conversational tasks since providing accurate responses that avoid hallucinations and adhere to known facts is often a minimum requirement.

\removedcontent{Persona-based models customize responses using auxiliary data that can represent the values, beliefs, characteristics, and/or history of a specific user.}
\newcontent{Persona-based models~\cite{liu2023pcpe,humeau2019poly,zhang2018personalizing,li2016persona} customize responses using auxiliary data that can represent the values, beliefs, characteristics, and/or history of a specific user.}
Models without persona information often fail to align responses with the user's preferences, which can be problematic when personalization is an expectation~\cite{liu2022persona}.
Many pre-fusion-based methods~\cite{humeau2019poly} concatenate personas with input queries as long text inputs to the model. 
These approaches have been shown to be sub-optimal because they process heterogeneous data from multiple sources in a homogeneous way, which might ignore the critical source-specific signals and increase the learning complexity of the model. 
% In other words, both the persona and the query are treated as text, but they carry different types of information from different sources. Furthermore, these methods cannot be easily extended to accommodate multi-modal data, such as attribute-based or event-based persona data. 
Liu~\etal~\cite{liu2023pcpe} recently proposed a retrieval model in a post-fusion framework that handles heterogeneous persona and query data in specialized encoding streams.

\subsection{Persona and Knowledge-based Models}
\label{sec:related:persona-knowledge}
The above methods are limited to a single source of auxiliary data, while many real applications contain multiple auxiliary sources that can provide valuable perspectives of the conversation.
Jang~\etal~\cite{jang2022call} proposed a pre-fusion-based method, \pkfocus, that identifies relevant persona and knowledge data based on a single highly summarized context embedding. This type of pre-fusion approach can ignore important signals in low-level word interactions. Consequently, it is unable to effectively extract useful information for response generation.

Liu~\etal~\cite{liu2023context} used low-level word similarities in the \pkncli method, which leverages a variation of ColBERT~\cite{khattab2020colbert} similarity in grounding networks. However, ColBERT is designed particularly for one-to-many document retrieval problems, where there is only one query and many documents. \pkncli failed to address critical issues of ColBERT when used for many-to-many context retrieval when it involves multiple sources, each with multiple entries.
\removedcontent{
\textbf{First}, ColBERT exhibits a bias toward the length of the 'query.' While this bias is harmless when only a single query is involved, it can falsely favor longer—but potentially irrelevant—context entries in multi-source, multi-entry scenarios. 
\textbf{Second}, it is an asymmetric metric that finds ``one-way'' relevance from the query to the documents. It ignores the critical reverse information flow from the documents to the queries when multiple auxiliary sources are involved. 
\textbf{Third}, it considers all word pairs equally, which makes it sensitive to frequent but noisy words (e.g., function words such as ``a'' and ``the'') in the language and causes it to suppress subtle but critical relevance among meaningful notional words.}
\newcontent{
\textbf{First}, ColBERT has a bias based on query length. While this bias is harmless when only one query is involved, it can favor longer, but potentially irrelevant, context entries in multi-source, multi-entry scenarios.
\textbf{Second}, it is an asymmetric metric that finds ``one-way'' relevance from the query to the documents. It ignores the critical reverse relevance from the documents to the queries when multiple auxiliary sources are involved. 
\textbf{Third}, it considers all word pairs equally, which makes it sensitive to frequent but mostly irrelevant words and suppresses subtle yet critical relevance among meaningful notional words.}

%% file: sections/method.tex
\section{Method}
\label{sec:method}
% \begin{figure}[h]
%     \vspace{-25pt}
%     \centering
%     \begin{minipage}[b]{.49\textwidth}
%         \includegraphics[width=\linewidth]{figures/CoMAC.pdf}
%         \vspace{-15pt}
%         \caption{\ourmethod Method Overview}
%         \label{fig:our-approach}
%     \end{minipage}
%     %
%     \begin{minipage}[b]{.49\textwidth}
%         \centering
%         \includegraphics[width=\linewidth]{figures/SSNColBERT.pdf}
%         \caption{Demonstration of $\ssncolbertsim$}
%         \label{fig:ssncolbert}
%     \end{minipage}
%     \vspace{-18pt}
% \end{figure}

%In this paper, we consider the problem of personalized and knowledge-grounded conversation response generation. 
In this paper, we consider a response generation problem grounded by auxiliary persona and knowledge data. 
Given 
a set of user persona entries $P$=$\{P_1, \dots, P_{N_p}\}$,
a set of knowledge entries related to the conversation $K$=$\{K_1, \dots, K_{N_k}\}$,
and the conversation history or utterance $U$ (concluding with a question), we aim to train an end-to-end model that can a response or answer, $\answer$, that is both personalized to the user and grounded in relevant knowledge.
$N_p$/$N_k$ are the numbers of entries in $P$/$K$. 
A persona entry, $P_i$, describes specific personal information about the user, while a knowledge entry, $K_j$, pertains to the conversation topic. Importantly, $U$, $\answer$, and the $P_i$ and $K_j$ entries are all in textual format.

% \begin{figure}[!t]
%     \centering
%     \includegraphics[width=0.7\linewidth]{figures/SSNCLI.pdf}
%     \caption{Overview of \ourmethod Architecture}
%     \vspace{-15pt}
%     \label{fig:our-approach}
% \end{figure}

\begin{wrapfigure}{r}{7cm}
        \centering
        \vspace{-20pt}
        \includegraphics[width=\linewidth]{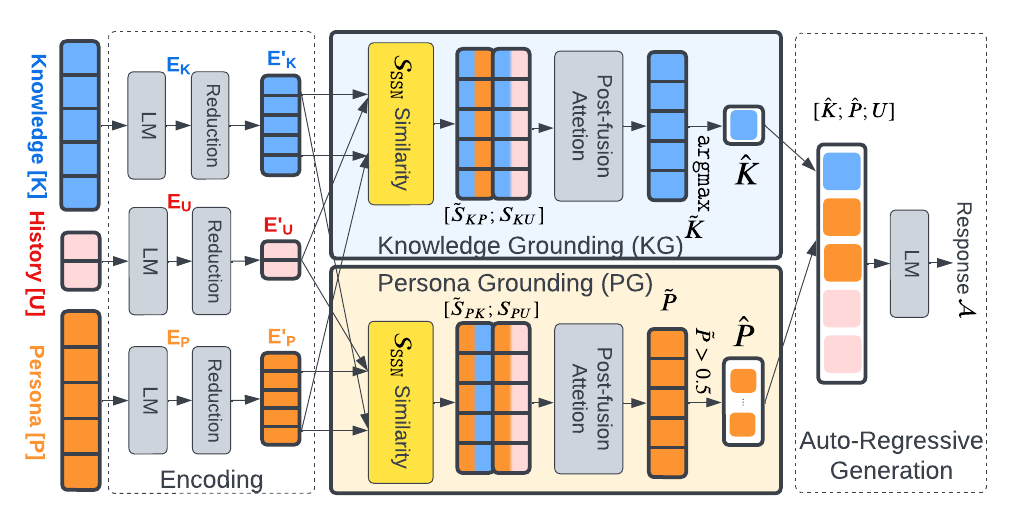}
        \vspace{-20pt}
        \caption{\ourmethod Method Overview}
        \vspace{-25pt}
        \label{fig:our-approach}
\end{wrapfigure}
\removedcontent{
Figure~\ref{fig:our-approach} demonstrates the \ourmethod method.
On a high-level, \ourmethod has three stages.
In the \textit{Input Encoding} stage (Section~\ref{sec:method:input}), the utterance $U$ and auxiliary data $P$/$K$ are encoded in specialized streams.
In the \textit{Post-Fusion Grounding} stage (Section~\ref{sec:method:grounding}), relevant subsets of persona entries $\hat{P} \in P$ and knowledge entries $\hat{K} \in K$ are identified in two separate post-fusion-based persona grounding (PG) and knowledge grounding (KG) networks by using the novel $\ssncolbertsim$ metric (Section~\ref{sec:method:sim}).
In the \textit{Generation} stage (Section~\ref{sec:method:generation}), the utterance, $U$, is supplemented with $\hat{P}$ and $\hat{K}$ to generate the response, $a$, using a fine-tuned language model $LM$ (i.e., $LM([\hat{P}; \hat{K}; U]) \rightarrow a$).
We emphasize the \textit{Post-Fusion Grounding Stage} as it highlights the main contributions of this paper.
Note that the \ourmethod method is easily extendable to other applications with additional auxiliary data sources by adopting new encoding streams and grounding networks.}
\newcontent{
Figure~\ref{fig:our-approach} demonstrates the three stages of the \ourmethod method.
In the \textbf{Encoding Stage} (Section~\ref{sec:method:input}), all inputs $U$/$P$/$K$ are embedded by a fine-tuned language model, $LM$.
In the \textbf{Post-Fusion Grounding Stage} (Section~\ref{sec:method:grounding}), two subsets of relevant persona and knowledge entries $\hat{P}$/$\hat{K}$ are selected in two separate post-fusion-based grounding networks, PG and KG, using the novel metric $\ssncolbertsim$ (Section~\ref{sec:method:sim}).
In the \textbf{Generation Stage} (Section~\ref{sec:method:generation}), the utterance, $U$, is supplemented with $\hat{P}$ and $\hat{K}$ to generate the response, $\answer$, using $LM$ (i.e., $LM([\hat{P}; \hat{K}; U]) \rightarrow \answer$).}

\subsection{Input Embedding}
\label{sec:method:input}
\removedcontent{
The utterance, $U$, candidate persona entries, $P$, and candidate knowledge entries, $K$, are first embedded by $LM$, respectively, as
$E_U = LM(\text{U}) \in \real^{1 \times s_u \times d}$,
$E_P = LM(\text{P}) \in \real^{N_p \times s_p \times d}$, and
$E_K = LM(\text{K}) \in \real^{N_k \times s_k \times d}$,}
\newcontent{
The inputs, $U$, $P$, and $K$, are first embedded separately by $LM$ as
$E_U \in \real^{1 \times s_u \times d}$,
$E_P \in \real^{N_p \times s_p \times d}$, and
$E_K \in \real^{N_k \times s_k \times d}$,}
where $s_u$/$s_p$/$s_k$ are the padded sequence lengths (i.e., number of tokens) of the $U$/$P$/$K$ entries, and $d$ is the embedding size. 
Here, each source of data is encoded in a separate stream to preserve low-level authentic information of the corresponding source, instead of using a single highly summarized embedding for all sources as in~\cite{jang2022call}.

\begin{comment}
    Theoretically, \ourmethod could use any existing language embedding model that converts from the vocabulary space to language embedding vectors. GPT-2~\cite{radford2019gpt2} and BART~\cite{lewis2019bart} are used in the experiments of Jang~\etal~\cite{jang2022call} and Liu~\etal~\cite{liu2023context}. However, based on our experimental results, we found that GPT-2's decoder-only architecture is not suitable for the PG/KG task. Therefore, we only present results based on the BART model in this paper. More details on the choice of LM will be discussed in Section~\ref{sec:discussion:base-lm}.
\end{comment}

\subsection{Sparse, Symmetric, Normalized ColBERT Similarity}
\label{sec:method:sim}

\removedcontent{The first step in identifying the relevant entries for response generation is to create a metric that evaluates the relevance of an auxiliary entry (e.g., $P$) to the rest of the conversational context (e.g., $K$ and $U$). 
As preliminary information, we review the ColBERT similarity~\cite{khattab2020colbert} (denoted as $\colbertsim$).
$\colbertsim$ measures the similarity between a query, $x$, and a document, $y$, in document retrieval problems by their low-level token latent interaction, computed as 
$\colbertsim(x, y) = \sum_{x_i \in x} \max_{y_j \in y} E_{x_i} \cdot E_{y_j}^T$,
where $E_{x_i}$/$E_{y_j}$ are the token embeddings of the $i$-th/$j$-th token of $x$/$y$, respectively. 
$\colbertsim$ can be generalized to any two text entries, $x$ and $y$.}
\newcontent{Identifying the relevant auxiliary entries starts with a metric that evaluates the relevance of an entry (e.g., $P$) to the rest of the conversational context (e.g., $K$ and $U$).
We first review the ColBERT similarity $\colbertsim$~\cite{khattab2020colbert}. $\colbertsim$ measures the token-level latent interaction similarity between a query, $x$, and a document, $y$, in document retrieval, computed as 
$\colbertsim(x, y) = \sum_{x_i \in x} \max_{y_j \in y} E_{x_i} \cdot E_{y_j}^T$,
where $E_{x_i}$/$E_{y_j}$ are the embeddings of the $i$-th/$j$-th token of $x$/$y$, respectively. 
$\colbertsim$ can be generalized to any two texts, $x$ and $y$.}

In our problem, the vanilla $\colbertsim$ suffers from several issues mentioned in Section~\ref{sec:related:persona-knowledge}.
Here, we develop a novel metric, denoted as $\ssncolbertsim$, that introduces \textbf{normalization}, \textbf{symmetry} and \textbf{sparsity} in ColBERT-style latent interactions.

% \subsubsection{Normalized ColBERT Similarity}
% \label{sec:method:sim:ncolbert}
\textbf{Normalization }
To eliminate the dominating bias of longer but potentially less relevant $x$'s in a batch (e.g., longer vs. shorter paragraphs in the auxiliary knowledge base), $\colbertsim$ is normalized by $|x|$, the number of tokens in $x$. That is,

\vspace{-7pt}
\[
% \begin{equation}
% \label{eqn:ncolbert}
\begin{aligned}
    % \texttt{ColBERT}(x, y)  & = \sum_{x_i \in x} \max_{y_j \in y} E_{x_i} \cdot E_{y_j}^T, \\
    % \texttt{NColBERT}(x, y) & = \frac{1}{|x|} \texttt{ColBERT}(x, y),
    \ncolbertsim(x, y) = \frac{1}{|x|} \colbertsim(x, y) = \frac{1}{|x|} \sum_{x_i \in x} \max_{y_j \in y} E_{x_i} \cdot E_{y_j}^T.
\end{aligned}
% \end{equation}
\vspace{-5pt}
\]

%
% \subsubsection{Symmetric Normalized ColBERT Similarity}
% \label{sec:method:sim:sncolbert}
\textbf{Symmetry }
Both $\colbertsim$ and $\ncolbertsim$ are asymmetric metrics that exchange information from only $x$ to $y$. In a multi-source auxiliary setup, it is critical to allow information to flow bi-directionally among all sources. A simple yet effective approach is to combine the asymmetric similarities obtained in both directions, i.e., $\sncolbertsim(x, y) = \ncolbertsim(x, y) + \ncolbertsim(y, x)$.

% \subsubsection{Sparse Symmetric Normalized ColBERT Similarity}
% \label{sec:method:sim:ssncolbert}
\textbf{Sparsity }
Another common issue of existing ColBERT-style similarities is that, when measuring pairwise token similarity, certain frequent tokens (such as function words) can dominate overall similarities, diminishing the subtle but critical signals from informative tokens (like notional words). To mitigate the potential adverse effects of such language noise, we introduce sparsity and selectivity into the similarity measure, using only a subset of tokens chosen through specific sampling strategies, denoted as
% \begin{equation}
% \label{eqn:ssncolbert}
% \begin{aligned}
%     \texttt{SSNColBERT}(x, y) & = \texttt{SNColBERT}(\hat{x}, \hat{y}),
% \end{aligned}
% \end{equation}
$\ssncolbertsim(x, y) = \sncolbertsim(\hat{x}, \hat{y})$,
where $\hat{x}$/$\hat{y}$ are the tokens sampled from $x$/$y$, respectively, following the sampling strategies.

The main assumption of sparsity and selectivity is that frequent tokens are less informative than those appearing sparsely in the dataset. TF-IDF~\cite{salton1988term} is a common solution that fits the assumption. We propose a sampling strategy that selects tokens with highest TF-IDF weights for $\ssncolbertsim$, referred to as ``\ourmethod (TF-IDF)''. We denote the percentage of selected tokens as $\sncr$.
% Our experiments and conclusions are based mostly on ``\ourmethod (TF-IDF)''.
In Section~\ref{sec:results:sparcity}, we also discuss an alternative sampling strategy which learns the token weights by a feed-forward network using $E_U$/$E_P$/$E_K$, referred to as ``\ourmethod (FF)''.

% \begin{figure}[t!]
% \centering
% %
% \begin{subfigure}{0.49\linewidth}
%     \centering
%     \includegraphics[width=\linewidth]{figures/ColBERT.pdf}
%     \caption{Demonstration of $\texttt{ColBERT}$ }
%     \label{fig:colbert}
% \end{subfigure}
% %
% \begin{subfigure}{0.49\linewidth}
%     \centering
%     \includegraphics[width=\linewidth]{figures/SSNColBERT.pdf}
%     \caption{Demonstration of $\texttt{SSNColBERT}$ }
%     \label{fig:ssncolbert}
% \end{subfigure}
% \caption{Comparison of $\texttt{ColBERT}$ and $\texttt{SSNColBERT}$ }
% \vspace{-10pt}
% \label{fig:colbert_vs_ssncolbert}
% \end{figure}

%Figure~\ref{fig:colbert_vs_ssncolbert} shows the comparison between $\texttt{ColBERT}$ and $\texttt{SSNColBERT}$.
%
\begin{wrapfigure}{rb}{5.5cm}
    \centering
    \vspace{-25pt}
    \includegraphics[width=\linewidth]{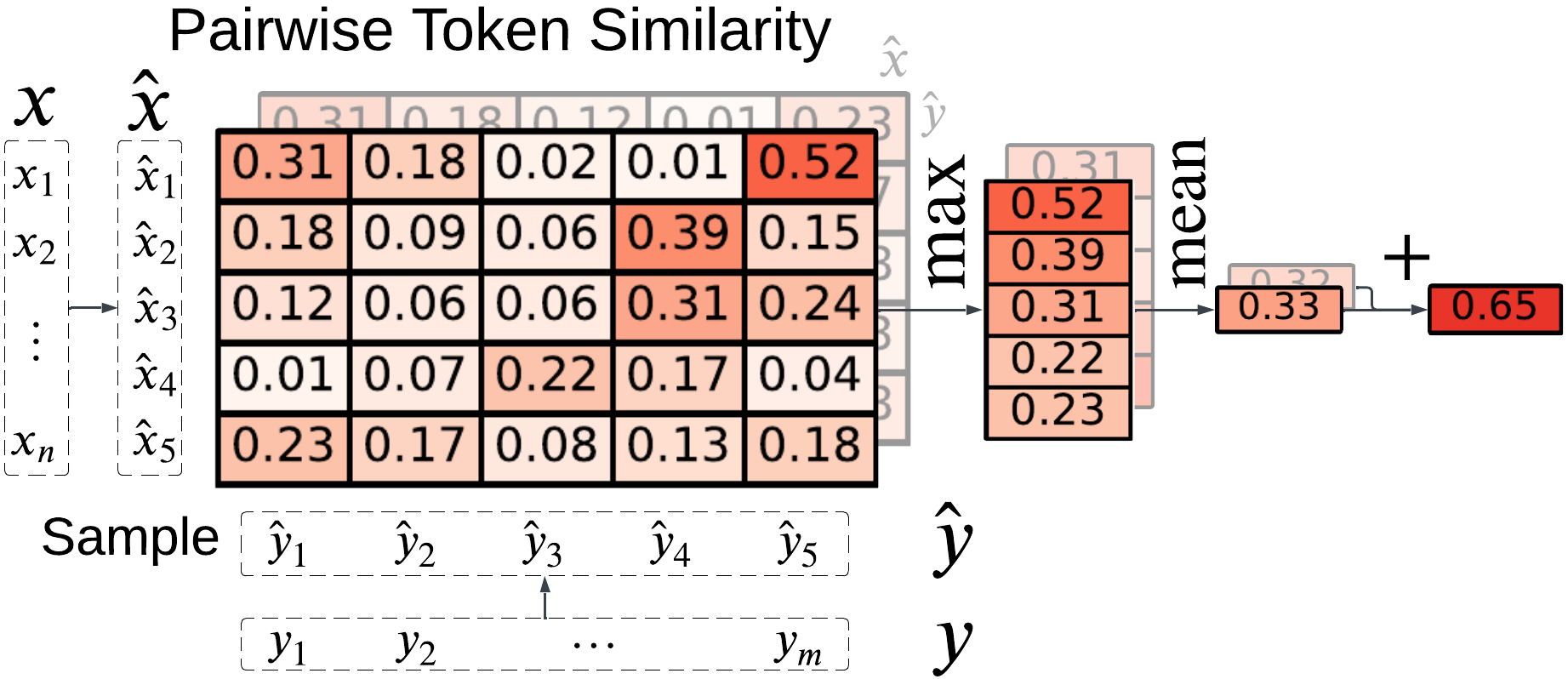}
    \vspace{-20pt}
    \caption{Demonstration of $\ssncolbertsim$}
    \vspace{-20pt}
    \label{fig:ssncolbert}
\end{wrapfigure}

\removedcontent{Figure~\ref{fig:ssncolbert} demonstrates the computation of $\ssncolbertsim$ similarity between two text entries $x$ and $y$ by using the token-level similarity of each pair of $(x_i, y_j)$. 
Calculating a similarity matrix for a large set of text entries $P$ and $K$ incurs an exponentially growing computational cost. To improve training and inference efficiency, instead of directly using $E_U$/$E_P$/$E_K$ for $\ssncolbertsim$, 
the embeddings are reduced to a lower dimension $d_0$ ($d_0 < d$) by a reduction layer $Re$, that is, $E'_U, E'_P, E'_K = Re(E_U), Re(E_P), Re(E_K)$.
% we employ a feed-forward layer (denoted as $Re$) to reduce the embeddings to a lower dimension $d_0$ ($d_0 < d$), that is, $E'_U, E'_P, E'_K = Re(E_U), Re(E_P), Re(E_K)$.
We set $d_0 = d/4$ in our experiments.}
\newcontent{Figure~\ref{fig:ssncolbert} demonstrates the computation of $\ssncolbertsim$ similarity by using token-level similarities.
Calculating a similarity matrix for a large set of text entries, $P$ and $K$, incurs an exponentially growing computational cost. 
To improve efficiency, we reduce $E_U/E_P/E_K \in \real^{d}$ to $E'_U, E'_P, E'_K \in \real^{d_0}$ ($d_0 = d/4$) by a dimension reduction layer prior to $\ssncolbertsim$.}

\subsection{Post-Fusion Grounding Networks}
\label{sec:method:grounding}
In this section, we aim to identify two subsets of relevant persona and knowledge entries, $\hat{P} \in P$, and knowledge entries, $\hat{K} \in K$, that are helpful for response generation in the PG and KG networks with $\ssncolbertsim$.

\textbf{Persona Grounding } In PG, we consider $P$ as the ``query'' set and $U$/$K$ as two distinct sets of ``documents.'' We construct two similarity matrices for: (I) persona and utterance entry pairs ($S^{PU}$), and (ii) persona and knowledge entry pairs ($S^{PK}$). The elements in $S^{PU}$ and $S^{PK}$ are calculated as follows:
\begin{equation}
\label{eqn:sim-matrices-p}
\begin{aligned}
    S^{PU}_{i,j} = \ssncolbertsim(E'_{P_i}, E'_{U_j}),\
    S^{PK}_{i,j} = \ssncolbertsim(E'_{P_i}, E'_{K_j}),  \\
\end{aligned}
\end{equation}
where $E'_{\cdot_i}$/$E'_{\cdot_j}$ is a reduced embedding of an entry token-by-token. We further calculate the similarities between each persona entry and the overall knowledge by taking the average entry-wise similarity, i.e.,  $\tilde{S}^{PK} = \frac{1}{N_k} \sum_{i=1}^{N_k} S^{PK}_{i}$.

\removedcontent{Each persona entry is scored by a post-fusion strategy that fuses the utterance relevance, $S^{PU}$, and the knowledge relevance, $\tilde{S}^{PK}$, in an attention layer with a sigmoid activation, i.e., 
$\tilde{P} = \sigma(w_1 \tilde{S}^{PK} + w_2 S^{PU} + b)$,
where $w_1$/$w_2$/$b$ are the attentions and bias, and $\tilde{P} \in \real^{N_p}$. A persona entry with $\tilde{P} > 0.5$ will be considered relevant, that is, $\hat{P} = P[\tilde{P} > 0.5]$.}
\newcontent{A post-fusion strategy scores each persona entry by fusing $S^{PU}$ and $\tilde{S}^{PK}$ in an attention layer with sigmoid activation, i.e., 
$\tilde{P} = \sigma(w_1 \tilde{S}^{PK} + w_2 S^{PU} + b)$,
where $w_1$/$w_2$/$b$ are the attentions and bias, and $\tilde{P} \in \real^{N_p}$.
Persona entries are considered relevant when $\hat{P} = P[\tilde{P} > 0.5]$.}

\textbf{Knowledge Grounding } 
\removedcontent{We use a similar idea to construct KG to identify the relevant knowledge entries. First, the knowledge-utterance similarity is computed as $S^{KU}_{i,j} = \ssncolbertsim(E'_{K_i}, E'_{U_j})$, and knowledge-persona similarity is simply $S^{KP} = (S^{PK})^T$ as $\ssncolbertsim$ is symmetric. 
The average knowledge-persona similarity is computed as $\tilde{S}^{KP} = \frac{1}{N_p} \sum_{i=1}^{N_p} S^{KP}_{i}$ to reduce the dominating effects from multiple persona entries. Then we fuse $S^{KU}$ and $\tilde{S}^{KP}$ with the post-fusion strategy, that is,
$\tilde{K} = \softmax(w_1 \tilde{S}^{KP} + w_2 S^{KU} + b)$.
The knowledge candidate with the highest $\tilde{K}$ score is estimated as relevant, 
% A relevant knowledge entry $\hat{K}$ is estimated by the candidate with the highest $\tilde{K}$ score,
i.e., $\hat{K} = K[\argmax(\tilde{K})]$.
Here \softmax and \argmax are used, following the dataset assumption in~\cite{jang2022call} that a response is based on only one knowledge entry, while multiple persona entries could contribute to it.}
\newcontent{KG is similarly constructed by calculating the relevance scores of each knowledge entry with the utterance and persona, $S^{KU}_{i,j}$ and $\tilde{S}^{KP}$, respectively, followed by the post-fusion strategy $\tilde{K} = \softmax(w_1 \tilde{S}^{KP} + w_2 S^{KU} + b)$.
The knowledge candidate with the highest $\tilde{K}$ score is considered relevant, 
% A relevant knowledge entry $\hat{K}$ is estimated by the candidate with the highest $\tilde{K}$ score,
i.e., $\hat{K} = K[\argmax(\tilde{K})]$.
Here \softmax and \argmax are used, following the dataset assumption~\cite{jang2022call} that a response is based on only one knowledge entry, while multiple persona entries could contribute to it.}

\subsection{Response Generation}
\label{sec:method:generation}

With estimated relevant persona and knowledge entries ($\hat{P}$ and $\hat{K}$), a new input $U^*$ is formed by combining $\hat{P}$, $\hat{K}$ and the original input $U$ for language model $LM$, i.e., $E_{U^*}$=$LM(U^*)$=$LM([\hat{K}; \hat{P}; U])$. A probability distribution $p_r$ over the vocabulary space is estimated based on $E_{U^*}$ by a projection layer.

\textbf{For inference}, $LM$ follows the auto-regressive framework~\cite{hoogeboom2021autoregressive}  to generate a sequence of tokens as the response.
\textbf{For training}, we train the model with loss function
\mbox{$L^{SSN} = \alpha L_{K} + \beta L^*_{P} + \gamma L_{M}$},
where $L_{K}$ is the cross-entropy loss over $\tilde{K}$ for knowledge grounding, 
$L^*_{P}$ is the cross-entropy loss over $\tilde{P}$ for persona grounding, 
$L_{M}$ is  the cross-entropy loss over $p_r$, and $\alpha$/$\beta$/$\gamma$ are the weighting hyper-parameters.
The loss term $L^*_{P}$ addresses an extreme imbalance issue overlooked by the baselines~\cite{jang2022call,liu2023context} that nearly 87\% of persona entries are irrelevant. 
First, we apply weights $w^*$/$(1-w^*)$ to positive/negative persona labels, respectively, to emphasize the importance of relevant persona entries.
Second, we randomly discard $L^*_{P}$ with probability of $p^*$ if an utterance has all negative persona labels. 
However, $L_{K}$ and $L_{M}$ are not affected by the persona labels.

%
% This loss function differs in terms of $L^*_{P}$ from the baseline methods~\cite{jang2022call}\cite{liu2023context} due to the extreme imbalance in the persona labels (nearly 87\% of persona entries are irrelevant). We combine two strategies to address the label imbalance issue.
% %
% With \textit{label weighting}, we apply weights of $w^*$/$(1-w^*)$ to positive and negative persona labels, respectively, to magnify the importance of relevant persona entries. 
% %
% With \textit{label sampling}, if an utterance has all negative persona labels, there is a probability of $p^*$ that $L^*_{P}$ will be used for training, and $(1-p^*)$ that $L^*_{P}$ will be discarded. However, $L_{K}$ and $L_{L}$ will always be kept for that utterance, regardless of the persona labels. 
% In our experiments, we set $w^*=0.9$ and $p^*=0.1$ as this combination roughly matches the distribution of the persona labels in the dataset and outperforms other combinations. 

% For inference, $LM$ follows the auto-regressive framework~\cite{hoogeboom2021autoregressive}  to generate a sequence of tokens as the response.

% Editing finished by Raju, 7/14/24

%% file: sections/experiments.tex
\section{Experiments}
\label{sec:experiments}

\subsection{Dataset}
\label{sec:experiments:dataset}

\input{tables/dataset}

We train \ourmethod models on the FoCus dataset~\cite{jang2022call}, which is a conversational dataset powered by persona and knowledge. Each conversation includes five text-based persona entries describing the user's personal profile, such as demographic data or hobbies (e.g., ``I find heritage-listed buildings interesting''), and several knowledge entries extracted from a Wikipedia page about a landmark (e.g., ``Thorps Building is a heritage-listed commercial building at Macrossan Street...''). For each round, zero or more persona entries are relevant to the conversation, while only one knowledge entry is relevant. Table~\ref{tab:dataset} shows some basic statistics of dataset. For more detailed information and examples, please refer to~\cite{jang2022call}.

% We consider the FoCus dataset~\cite{jang2022call} as a suitable dataset for training \ourmethod. The FoCus dataset was generated by crowd-sourced workers who were asked to conduct conversations based on a given set of candidate persona and knowledge entries. Each conversation includes five text-based persona entries describing the user's personal profile such as demogrpahic data or hobbies (e.g., \textcolor{red}{``I find heritage-listed buildings interesting''}) and ground truth knowledge entries about a conversation topic represented by paragraphs extracted from a Wikipedia page about a landmark (e.g., \textcolor{red}{``Thorps Building is a heritage-listed commercial building at Macrossan Street...''}). The dataset also provides labels indicating which persona/knowledge entries were used in each round (utterance).

% The FoCus dataset consists of 13,484 conversations about 6,075 landmarks (topics), divided into 12,484 training conversations and 1,000 validation conversations. Each conversation averages 5.63 rounds. While all answers are knowledge-based, they do not necessarily incorporate personas. The training dataset comprises 37,488 knowledge-only answers and 32,855 persona-knowledge answers, with a similar ratio maintained in the validation dataset (3,007 vs. 2,630). For detailed dataset statistics and examples, please refer to the original paper.

\subsection{Experimental Setup}
\label{sec:experiments:setup}

We compare \ourmethod with two baseline methods, \pkfocus~\cite{jang2022call} and \pkncli~\cite{liu2023context}. The source code of \ourmethod is available online~\footnote{\url{https://github.com/jliu-v/CoMAC}}. All models are initialized with pre-trained BART~\cite{lewis2019bart} weights, trained on NVIDIA RTX2080-Ti GPUs over the training set for two epochs and evaluated on the validation set.
\newcontent{For TF-IDF, the IDF weights are pre-computed offline for efficiency, and the TF and final TF-IDF weights are computed on the fly.}
It is worth noting that, theoretically, \ourmethod could use any existing language model $LM$. GPT-2~\cite{radford2019gpt2} is sub-optimal compared to BART (encoder-decoder architecture) for all methods, as GPT-2's decoder-only architecture is not suitable for the PG/KG tasks which require advanced capabilities on comprehension and summarization.

\textbf{Hyper-parameters}
Besides the default setup of $\alpha/\beta/\gamma$=1/1/10 in~\cite{jang2022call}, we conduct additional hyper-parameter search, limiting $\alpha$+$\beta$+$\gamma$=10 to avoid arbitrary optimization solutions. 
\removedcontent{For TF-IDF, inverse document frequency (IDF) weights are pre-computed using Scikit-Learn~\footnote{\url{https://scikit-learn.org}}. Term frequency (TF) and final TF-IDF weights are computed on the fly.}
In our experiments, the optimal values of other hyper-parameters are $w^*$=$0.9$, $p^*$=$0.1$ and $\sncr$=$0.35$. The $w^*$/$p^*$ values roughly match the persona label distribution in the dataset.

\textbf{Evaluation Metrics}
We evaluate \ourmethod in terms of both the quality of generated responses and the grounding accuracies. 
For response quality, we use the perplexity (\textbf{PPL}), \textbf{ROUGE} and \textbf{BLEU}, and \textbf{F1} scores. 
For grounding, we use \textbf{PG}/\textbf{KG} accuracies. Due to the imbalanced persona label distribution, we also report F1, precision, and recall for PG (\textbf{PG-F1}/\textbf{PG-PR}/\textbf{PG-RC}).
All reported metrics are based on the validation dataset. 
Lower PPL values indicate better performance, while the other metrics are the higher the better.

%% file: tables/dataset.tex
\begin{wraptable}{r}{4.5cm}
\vspace{-25pt}
\centering
\captionsetup{font=scriptsize}
\caption{FoCus Dataset Overview}
\vspace{-10pt}
\resizebox{\linewidth}{!}{%
\begin{tabular}{lrr} 
\toprule
                              & Train.    & Valid. \\
\midrule
\# Dialogs                    & 12484       & 1,000                     \\
\# Average Rounds             & 5.63        & 5.64                      \\
Avg. Length of Human's Utt.   & 40.70       & 40.21                     \\
Avg. Length of Machine's Utt.{ } & 138.16      & 138.60                    \\
\# Knowledge-only Answer      & 37,488      & 3,007                     \\
\# Persona-Knowledge Answer   & 32,855      & 2,630                     \\
\# Landmarks                  & 5,152       & 923                     \\
\bottomrule             
\end{tabular}
}
\label{tab:dataset}
\vspace{-20pt}
\end{wraptable}

%% file: sections/results.tex
\section{Results}
\label{sec:results}

\removedcontent{
\subsection{Overall Performance}
\label{sec:results:overall}
}

%TODO - Junfeng this sentance is not clear. See if the rewrite works.
%Table~\ref{table:performance:compare} presents the comparison of the best performing \ourmethod model with the baseline \pkfocus and \pkncli, as well as the \texttt{SNColBERT} (\ourmethod without sparse token selection).
%
Table~\ref{table:performance:compare} compares the experimental results of \ourmethod models with the baselines. 
The comparison demonstrates that our method, \ourmethod, surpasses both baselines in terms of both language generation and KG/PG qualities. Specifically, \ourmethod outperforms the best baseline \pkncli by 5.26\%, 9.31\%, 7.84\% and 11.64\% in F1, ROUGE-L, BLEU, and PPL scores, respectively. It also enhanced PG accuracy (PG) and F1 (PG-F1) by 45.75\% and 27.41\% , respectively, and KG accuracy (KG) by 6.76\%. \ourmethod outperforms \pkfocus  even more significantly.
This comparison indicates on a high level that \ourmethod can better leverage the sparsity and symmetry to identify the relevancy among various conversational contexts, which further assists language modeling and response generation.
% The superiority of \ourmethod indicates that the sparsity and symmetricality in \texttt{SSNColBERT} can help extract subtle discriminative signals from the text tokens and better capture the supplemental two-way signals from different auxiliary data sources. \ourmethod can better leverage this sparsity and symmetry to identify the relevancy among various conversational contexts, which further assists language modeling and response generation.
% 
% This indicates the superiority of our method (\ourmethod) over the baseline methods. The sparsity and symmetry of the entry similarity measure help extract discriminative signals from the text tokens and better capture the supplemental two-way signals from different text entries. \ourmethod can leverage this sparsity and symmetry to better identify the relevancy among various text-based conversational contexts, which further assists language modeling and response generation.

\input{tables/best_performance}

\subsection{Symmetric Information Flow}
\label{sec:results:symetric}
Asymmetric similarities like $\colbertsim$ or $\ncolbertsim$ learn one-way information flow  \textit{from} the ``queries'' (e.g., $P$ in PG network) \textit{to} the ``documents'' (e.g., $U$/$K$ as the context in PG network). They are not suitable for the purpose of the grounding networks in conversational problems, where all sources of inputs (whether ``query'' or ``documents'') will contribute to the final output.

\removedcontent{To support our claim that the supplemental two-way signals are critical in the problem with multiple auxiliary data, 
% we conduct experiments to compare the \pksncli method, using two-way \texttt{SNColBERT} similarity without token sampling strategy, against the base \pkncli, using one-way \texttt{NColBERT} similarity. 
we conduct experiments to compare the \ourmethod method with the two-way $\sncolbertsim$ (no sparsity, denoted as \pksncli) against the baseline \pkncli with the one-way $\ncolbertsim$. 
Table~\ref{table:performance:compare} shows that \pksncli has significant improvements over the baseline \pkncli on PG accuracy (51.82\% vs 44.75\%), and provides minor improvements on KG accuracy (90.42\% vs 90.25\%) and other language generation metrics.
For the PG task with extremely imbalanced and sparse data, it is especially important to exploit symmetry and utilize other contextual information to identify relevant persona entries.
For the KG task, it benefits less from symmetric text similarity, primarily because most contexts in the dataset are highly correlated with knowledge entries. For example, most questions are related to the knowledge of the conversation topic, making them more similar to traditional document retrieval tasks. Therefore, the asymmetric $\ncolbertsim$ similarity already captured most of the useful signals for KG, and the symmetric $\sncolbertsim$ similarity is able to provide marginal improvements.}
\newcontent{To demonstrate that the supplemental two-way signals are critical among multiple auxiliary data, we compare the \ourmethod method with the two-way $\sncolbertsim$ (no sparsity, denoted as \pksncli) against the baseline \pkncli with the one-way $\ncolbertsim$. 
Table~\ref{table:performance:compare} shows that \pksncli has significant improvements over the baseline \pkncli on PG accuracy (51.82\% vs 44.75\%), and provides minor improvements on KG accuracy (90.42\% vs 90.25\%) and other language generation metrics.
For PG tasks with extremely imbalanced and sparse data, it is especially important to exploit symmetry and leverage other context data.
The KG task benefits less from the symmetry, primarily because most contexts in the dataset are highly correlated with knowledge entries. Most questions are related to information from the knowledge, making them more similar to traditional document retrieval tasks. Therefore, the asymmetric $\ncolbertsim$ already captured most of the useful signal, and the symmetric $\sncolbertsim$ only provides marginal improvement.}

\input{tables/token_sample}

\subsection{Sparse Token Sampling}
\label{sec:results:sparcity}
The baselines failed to reduce the extra noise brought into the grounding networks from less informative tokens. Such noise not only dilutes the overall information in texts but could also obscure the importance of meaningful words. We conduct two sets of \ourmethod experiments with two different strategies (TF-IDF and FF) to sample tokens in the PG/KG networks.

Table~\ref{table:performance:compare} shows that both strategies can significantly improve the PG/KG and generation performance over the two baselines and \pksncli. TF-IDF/FF improved PG over the best baseline \pkncli by 45.74\%/62.46\% (65.22/72.70 vs 44.75), PG-F1 by 27.41\%/43.47\% and KG by 6.76\%/2.47\%, respectively. They also improved the generation quality, e.g., F1 by 5.26\%/3.51\%.
\newcontent{
Table~\ref{table:token-sample} shows an example of the tokens sampled by the \ourmethod (TF-IDF).
For efficiency, \ourmethod outperforms the baseline by 9.77\% (2872 vs 3183 sec.) in terms of inference time. In real applications, further efficiency and scalability can be achieved through pre-sampling of persona and knowledge fully offline.}

An interesting observation is that TF-IDF is better for KG and generation performances, while FF is better for PG performance. This is probably because many conversations in the dataset are highly correlated with knowledge, and personas do not have as much influence as knowledge on the responses, as observed in~\cite{liu2023context}. The FF strategy is a fully trained layer directly driven by the $L^*_{P}$ loss of the PG network. In contrast, the token distribution and contribution to the PG training target is disconnected. Therefore, FF performs well on PG accuracy. However, this doesn't make TF-IDF a sub-optimal choice overall, as it is directly tied to the token distributions in the dataset, which can preserve rare but informative tokens in the grounding stage for response generation later.

\subsection{\newcontent{Human Evaluation}}
\label{sec:results:human-evaluation}

\newcontent{
To evaluate the quality of generated responses, we randomly selected 75 conversations (408 rounds) from the validation set for human evaluation. We presented to human volunteers the original utterance, ground truth and predicted knowledge, as well as the persona entries, reference response and two predicted responses from \ourmethod and \pkfocus (denoted as $\answercomac$ and $\answerfocus$, respectively).
In order to compare the performance of \ourmethod with the baseline, each volunteer was asked to answer three questions: 
\textbf{Q1}. do they prefer $\answercomac$ or $\answerfocus$,
\textbf{Q2}. did $\answercomac$ successfully address the original question, and
\textbf{Q3}. did $\answerfocus$ successfully address the original question. 
For \textbf{Q1}, 46.81\% of $\answercomac$ were rated as better than $\answerfocus$, 42.40\% were similar to $\answerfocus$, and only 10.78\% were worse than $\answerfocus$. 
For \textbf{Q2} and \textbf{Q3}, 90.44\% and 75.98\% of $\answercomac$ and $\answerfocus$, respectively, addressed the original question. 
These results demonstrate that \ourmethod is better than \pkfocus at addressing questions and producing higher quality responses.}

%% file: tables/best_performance.tex
\begin{table*}[t]
\centering
\vspace{-10pt}
\captionsetup{font=footnotesize}
\caption{Summary of Performance Comparison}
\label{table:performance:compare}
\resizebox{\columnwidth}{!}{%
\begin{threeparttable}
%\begin{tabular}{l|rrrrrrrrrrr} 
\bgroup
\def\arraystretch{1}%
\begin{tabular}{
  @{\hspace{0pt}}l@{\hspace{5pt}}|
  @{\hspace{5pt}}r@{\hspace{5pt}}
  @{\hspace{0pt}}r@{\hspace{5pt}}
  @{\hspace{0pt}}r@{\hspace{5pt}}
  @{\hspace{0pt}}r@{\hspace{5pt}}
  @{\hspace{0pt}}r@{\hspace{5pt}}
  @{\hspace{0pt}}r@{\hspace{5pt}}
  @{\hspace{0pt}}r@{\hspace{5pt}}
  @{\hspace{0pt}}r@{\hspace{5pt}}
  @{\hspace{0pt}}r@{\hspace{5pt}}
  @{\hspace{0pt}}r@{\hspace{5pt}}
  @{\hspace{0pt}}r@{\hspace{5pt}}
}
\toprule
Method          & F1   & ROUGE1 & ROUGE2 & ROUGEL & BLEU & PPL$^{\downarrow}$   & PG(\%)  & PG-F1(\%) & PG-PR(\%) & PG-RC(\%) & KG(\%)  \\
\midrule
\pkfocusO    
                & 0.291 & 0.353 & 0.186 & 0.311 & 11.364 & 25.23  & \st{86.70} &     - &     - &  0.00 & 68.61 \\      
\pkfocus        
                & 0.304 & 0.377 & 0.214 & 0.335 & 12.966 & 11.863 & 13.40 & 23.46 & 13.30 & \bf{99.70} & 70.76 \\
\midrule
\pkncliO     
                & 0.317 & 0.382 & 0.213 & 0.337 & 12.882 & 13.17  & \st{86.69} &     - & -     &  0.00 & 89.61 \\
\pkncli         
                & 0.342 & 0.409 & 0.241 & 0.361 & 14.622 &  9.021 & 44.75 & 30.46 & 18.30 & 90.90 & 90.25 \\
\midrule
\pksncli
                & 0.348 & 0.410 & 0.243 & 0.361 & 14.632 &    9.19 & 51.82 & 32.60 & 20.05 & 87.73 & 90.42 \\
\midrule
\ourmethod (FF)
                & 0.354 & 0.420 & 0.254 & 0.372 & 15.386 &    9.07 & \bf{72.70} & \bf{43.70} & \bf{30.11} & 79.68 & 92.48 \\
\ourmethod (TF-IDF)$^*$
                & \bf{0.360} & \bf{0.429} & \bf{0.263} & \bf{0.381} & \bf{15.768} &    \bf{7.97} & 65.22 & 38.81 & 25.33 & 82.93 & \bf{96.35} \\
Improvement
                & 5.26\% & 4.89\% & 9.13\% & 5.54\% & 7.84\% & 11.64\% & 45.74\% & 27.41\% & 38.42\% & -8.77\% &  6.76\% \\

\bottomrule
\end{tabular}

\begin{tablenotes}[flushleft]
    \setlength\labelsep{0pt}
    \footnotesize
    \item 
    \pkfocusO and \pkncliO represent the baseline methods from the original publications where $w^*$ and $p^*$ were not used. PG performance of \pkfocusO and \pkncliO are discarded as due to over-fitting issue mentioned in Section~\ref{sec:method:generation}.
    \pkncli is equivalent to \ourmethod with only $\ncolbertsim$. 
    \pksncli is \ourmethod with only $\sncolbertsim$ (no token sampling).
    Metrics with ``$^{\downarrow}$'' mean that lower values are more desirable. 
    Values in \textbf{bold} represent the best performance of the corresponding metric among all methods from both default setup and hyper-parameter search experiments. ``Improvement'' measures the percentage improvement of the best \ourmethod model (denoted by ``$^{*}$'') over the best baseline \pkncli. 
\end{tablenotes}
\egroup 
\end{threeparttable}
}
\vspace{-20pt}
\end{table*}

%% file: tables/token_sample.tex
\begin{table}
\centering
\vspace{-25pt}
\captionsetup{font=footnotesize}
\caption{\newcontent{Example of Tokens Sampled by \ourmethod}}
\label{table:token-sample}
\resizebox{0.95\linewidth}{!}{%
\begin{threeparttable}
\bgroup
\def\arraystretch{1}%
\begin{tabular}{
    @{\hspace{0pt}}p{2.2cm}@{\hspace{5pt}}|
    @{\hspace{5pt}}l@{\hspace{5pt}}
}
\toprule

Persona ($\hat{P}$) & I \tokenhl{hope} to \tokenhl{move} to \tokenhl{Adelaide} this year . \\
\midrule

Knowledge ($\hat{K}$) & 
\begin{minipage}[t]{\columnwidth}%
The National \tokenhl{War} \tokenhl{Memorial} is a \tokenhl{monument} on the north \tokenhl{edge} of the city centre of \tokenhl{Adelaide} , South \tokenhl{Australia} , \tokenhl{commemorating} those who \tokenhl{served} in the \tokenhl{First} World \tokenhl{War} .
\end{minipage}
\\
\midrule

Utterance ($U$)& \tokenhl{Where} is this \tokenhl{memorial} ? \\
\midrule

Response ($\answer$) & 
\begin{minipage}[t]{\columnwidth}%
This memorial is located on the north edge of the city centre of Adelaide , South Australia , where you hope to move to this year . 
\end{minipage}
\\

\bottomrule
\end{tabular}

\begin{tablenotes}[flushleft]
    \setlength\labelsep{0pt}
    \footnotesize
    \item 
    \tokenhl{Tokens} with underscores and bold fonts are the sampled tokens from the original full text. 
\end{tablenotes}
\egroup 
\end{threeparttable}
}
\vspace{-30pt}
\end{table}

%% file: sections/ablation.tex
\section{Ablation Study}
\label{sec:ablation}

Previous work~\cite{liu2023context} showed that, among the three sub-tasks (KG, PG, LM), good learning on one can benefit the other two. However, the optimization objective prioritizes one task at the expense of the others when assigning weights. An ablation study is necessary to investigate how each hyper-parameter affects \ourmethod’s performance and find the optimal balance. In this study, we test the model’s sensitivity to $\alpha$/$\beta$/$\lambda$/$\sncr$. Except for the hyper-parameter being tested, other values are fixed to the optimal combination from the main experiments, that is, $\alpha$/$\beta$/$\lambda$/$\sncr$=1/1/10/0.35, TF-IDF for sampling, and $w^*$/$p^*$=0.9/0.1 for $L^*_{P}$ loss.
% we set others values with the default of $\alpha$/$\beta$/$\lambda$/$\sncr$=1/1/10/0.35, as this is the optimal combination from our experiments. We use TF-IDF for token sampling and $w^*$/$p^*$=0.9/0.1 for $L^*_{P}$ loss.

% Previous work~\cite{jang2022call}\cite{liu2023context} has shown that, among the three sub-tasks (KG, PG, $LM$) of the main response generation task, good learning on one task can benefit the other two. However, the optimization objective of the \ourmethod method (Equation~\ref{eqn:loss}), prioritizes one task at the expense of the others when assigning weights. Since the three sub-tasks both assist and compete with each other, maintaining a balance is important. In this section, we conduct an ablation study to find the optimal balance and investigate how each hyper-parameter affects \ourmethod’s performance. Specifically, we test the model’s sensitivity to the weights of the KG network ($\alpha$), PG network ($\beta$), $LM$ modeling ($\lambda$), and the sampling rate for the \texttt{SSNColBERT} similarity ($\sncr$). In these experiments, except for the hyper-parameter being tested, we set other hyper-parameters with the default values of $\alpha$=1, $\beta$=1, $\lambda$=10 and $\sncr$=0.35, as these are the optimal hyper-parameters from our experiments reported in Table~\ref{table:performance:compare}. We use TF-IDF strategy for token sampling and use $w^*=0.9$ and $p^*=0.1$ for the training loss.

% \subsection{Knowledge Grounding Weight: $\alpha$}
% \label{sec:ablation:kg}
% % Weight on KG.
\textbf{Knowledge Grounding ($\boldsymbol{\alpha}$) } 
Figure~\ref{fig:alpha} shows how $\alpha$ affects the performance of \ourmethod. 
When $\alpha$=0, the model performs poorly in KG and response generation, because the model fails to learn the KG network, and irrelevant knowledge entries are selected for generation. When $\alpha$=1, the performance improves significantly due to learning knowledge entries relevant to the conversation. As $\alpha$ increases, it diminishes the learning objectives for the PG and LM. Without a well-learned LM, it becomes difficult to extract signals from text-based knowledge. Therefore KG and generation performance declines despite $\alpha$ increases.

\begin{figure}
    \vspace{-21pt}
    \centering
    \begin{minipage}{0.49\columnwidth}
        \centering
        \input{figures/perf_change_bart_k}
    \end{minipage}
    % \hfill
    \hspace{1pt}
    \begin{minipage}{0.49\columnwidth}
        \centering
        \input{figures/perf_change_bart_p}
    \end{minipage}
    \vspace{-22pt}
\end{figure}

% \subsection{Persona Grounding Weight: $\beta$}
% \label{sec:ablation:pg}
% Weight on PG.

\textbf{Persona Grounding ($\boldsymbol{\beta}$) } 
Figure~\ref{fig:beta} demonstrates how $\beta$ affects the PG and generation performance of the \ourmethod model. We observe that a non-zero $\beta$ value is critical for improving the performance of the PG and LM tasks, but too much emphasis on the PG task can be detrimental.

Fig.~\ref{fig:beta-pg} and~\ref{fig:beta-lm} show that the optimal $\beta$ values are 6/1 for PG/LM tasks, respectively. This aligns with the observation that conversations are less correlated with personas than with knowledge. 
A common example in the dataset is questions, $U$, like ``Where is this place?''. A positive-labeled persona entry matches the reference answer $\answer$ ``This is \texttt{[place]}, a place you like to visit''. However, the core part of $\answer$ (``\texttt{[place]}'') might already be extracted from the conversation context (e.g., knowledge or history). In these cases, the labeled persona entry provides little new insight for the rest of the answer. Consequently, higher $\beta$ leads to higher PG performance, but lower performance on KG and generation. However, when $\beta$ is very high ($\beta$>6), an insufficiently trained LM will no longer adequately support the learning of the PG network due to weights competition.

\begin{wrapfigure}{rb}{4.5cm}
\input{figures/perf_change_bart_l}
\end{wrapfigure}

% \subsection{Language Model Weight: $\gamma$}
% \label{sec:ablation:lm}

%
\textbf{Language Modeling ($\boldsymbol{\gamma}$) }
Figure~\ref{fig:gamma-lm} shows how $\gamma$ impacts on the LM generation.
We experiment with various $\gamma$ values between 1$\sim$20. 
The best performance is achieved with $\gamma$=10. 
When $\gamma$<10, generation performance increases with $\gamma$ as the LM is able to model the language better. 
With higher $\gamma$, competition among the three sub-tasks takes effect, suppressing the learning of PG and KG. As a result, the LM is unable to learn effectively without the assistance of persona/knowledge data from well-trained PG/KG networks.

% \subsection{Sparsity Sampling Rate for \texttt{SSNColBERT}: $\sncr$}
% \label{sec:ablation:sncr}

% \input{figures/perf_change_bart_sncr}

\begin{wraptable}{r}{5.5cm}
\input{tables/bart_perf_sncr}
\end{wraptable}
\textbf{Sampling Rate ($\boldsymbol{\sncr}$) }
Table~\ref{table:ablation:sncr} summarizes performance of \ourmethod with $\sncr$ values of  [0.25, 0.5, 0.75, 1.0], in addition to the 0.35 reported in the main experiments.
We notice that, as long as the sampling is in place ($\sncr$<1.0), $\sncr$ has a very minor effect on the overall performance, while $\sncr$=0.35 is slightly better than the other values in most evaluation metrics.
Without sampling ($\sncr$=1.0), equivalent to \pksncli, the model's performance drops significantly due to its inability to mitigate the impact of noisy tokens. This stresses the significance of the sampling strategy in preserving useful signals, but more importantly, in reducing noise in the data.

%% file: figures/perf_change_bart_k.tex
    \centering
    \begin{subfigure}[t]{0.49\linewidth}
        \centering
        \resizebox{1.05\linewidth}{!}{\input{figures/k_kg}}
        % \begin{gnuplot}[terminal=cairolatex, scale=0.32]
        % set lmargin at screen 0.1
        % set size ratio 0.7
        % set datafile separator comma
        % set xrange [0:9]
        % set xlabel '\huge $\alpha$'
        % set ylabel '\Large KG'
        % set key outside above
        % set key height 1.1
        % plot 'figures/data/bart_k.csv' u 4:($18/100) w linespoints title '\large KG' axis x1y1
        % \end{gnuplot}
        \vspace{-15pt}
        \captionsetup{font=scriptsize}
        \caption{KG}
        \label{fig:alpha-kg}
    \end{subfigure}
    %
    %\hfill
    \hspace{-3pt}
    \begin{subfigure}[t]{0.49\linewidth}
        \centering
        \resizebox{1.13\linewidth}{!}{\input{figures/k_gen}}
        % \begin{gnuplot}[terminal=cairolatex, scale=0.32]
        % set lmargin at screen 0.1
        % set size ratio 0.7
        % set datafile separator comma
        % set xrange [0:9]
        % set xlabel '\huge $\alpha$'
        % set ylabel '\Large F1 / ROGUE'
        % set y2tics
        % set y2label '\Large BLEU / PPL$^{\downarrow}$'
        % set key outside above
        % plot 'figures/data/bart_k.csv' u 4:9 w linespoints title '\large F1' axis x1y1, \
        %      'figures/data/bart_k.csv' u 4:10 w linespoints title '\large ROGUE1' axis x1y1, \
        %      'figures/data/bart_k.csv' u 4:11 w linespoints title '\large ROGUE2' axis x1y1, \
        %      'figures/data/bart_k.csv' u 4:12 w linespoints title '\large ROGUEL' axis x1y1, \
        %      'figures/data/bart_k.csv' u 4:13 w linespoints title '\large BLEU' axis x1y2, \
        %      'figures/data/bart_k.csv' u 4:19 w linespoints title '\large PPL$^{\downarrow}$' axis x1y2
        % \end{gnuplot}
        \vspace{-15pt}
        \captionsetup{font=scriptsize}
        \caption{Generation}
        \label{fig:alpha-lm}
    \end{subfigure}
    \vspace{-8pt}
    \captionsetup{font=footnotesize}
    \caption{Effectiveness of $\alpha$}
    \label{fig:alpha}

%% file: figures/perf_change_bart_p.tex
    \centering
    \begin{subfigure}[t]{0.45\linewidth}
        \centering
        \resizebox{1.23\linewidth}{!}{\input{figures/p_pg}}
        % \begin{gnuplot}[terminal=cairolatex, scale=0.32]
        %     set lmargin at screen 0.13
        %     set size ratio 0.7
        %     set datafile separator comma
        %     set xrange [0:9]
        %     set xlabel '\huge $\beta$'
        %     set ylabel '\Large PG Scores'
        %     set key outside above
        %     set key vertical maxrows 2
        %     set ytics nomirror
        %     set ytics (0, 0.2, 0.4, 0.6, 0.8, 1.0)
        %     plot 'figures/data/bart_p.csv' u 5:($14/100) w linespoints title '\large PG', \
        %          'figures/data/bart_p.csv' u 5:($16/100) w linespoints title '\large PG-PR', \
        %          'figures/data/bart_p.csv' u 5:($17/100) w linespoints title '\large PG-RC', \
        %          'figures/data/bart_p.csv' u 5:(2*($16*$17)/($16+$17)/100) w linespoints title 'PG-F1'
        % \end{gnuplot}
        \vspace{-15pt}
        \captionsetup{font=scriptsize}
        \caption{PG}
        \label{fig:beta-pg}
    \end{subfigure}
    %
    % \hfill
    \hspace{3pt}
    \begin{subfigure}[t]{0.45\linewidth}
        \centering
        \resizebox{1.23\linewidth}{!}{\input{figures/p_gen}}
        % \begin{gnuplot}[terminal=cairolatex, scale=0.32]
        %     set lmargin at screen 0.1
        %     set size ratio 0.7
        %     set datafile separator comma
        %     set xrange [0:9]
        %     set xlabel '\huge $\beta$'
        %     set ylabel '\Large F1 / ROGUE'
        %     set y2tics
        %     set y2label '\Large BLEU / PPL$^{\downarrow}$'
        %     set key outside above
        %     plot 'figures/data/bart_p.csv' u 5:9 w linespoints title '\large F1' axis x1y1, \
        %          'figures/data/bart_p.csv' u 5:10 w linespoints title '\large ROGUE1' axis x1y1, \
        %          'figures/data/bart_p.csv' u 5:11 w linespoints title '\large ROGUE2' axis x1y1, \
        %          'figures/data/bart_p.csv' u 5:12 w linespoints title '\large ROGUEL' axis x1y1, \
        %          'figures/data/bart_p.csv' u 5:13 w linespoints title '\large BLEU' axis x1y2, \
        %          'figures/data/bart_p.csv' u 5:19 w linespoints title '\large PPL$^{\downarrow}$' axis x1y2
        % \end{gnuplot}
        \vspace{-15pt}
        \captionsetup{font=scriptsize}
        \caption{Generation}
        \label{fig:beta-lm}
    \end{subfigure}
    \vspace{-8pt}
    \captionsetup{font=footnotesize}
    \caption{Effectiveness of $\beta$}
    \label{fig:beta}

%% file: figures/perf_change_bart_l.tex
    \vspace{-22pt}
    \centering
    
    \resizebox{\linewidth}{!}{\input{figures/l_gen}}
    % \begin{gnuplot}[terminal=cairolatex, scale=0.35]
        
    %     set size ratio 0.55
    %     set datafile separator comma
    %     set xlabel '\huge $\gamma$'
    %     set ylabel '\Large F1 / ROGUE'
    %     set y2tics
    %     set y2label '\Large BLEU / PPL$^{\downarrow}$'
    %     set key outside above
    %     set xrange [0:20]
    %     set yrange [0.19:0.46]
    %     plot 'figures/data/bart_l.csv' u 3:9 w linespoints title 'F1' axis x1y1, \
    %          'figures/data/bart_l.csv' u 3:10 w linespoints title 'ROGUE1' axis x1y1, \
    %          'figures/data/bart_l.csv' u 3:11 w linespoints title 'ROGUE2' axis x1y1, \
    %          'figures/data/bart_l.csv' u 3:12 w linespoints title 'ROGUEL' axis x1y1, \
    %          'figures/data/bart_l.csv' u 3:13 w linespoints title 'BLEU' axis x1y2, \
    %          'figures/data/bart_l.csv' u 3:19 w linespoints title 'PPL$^{\downarrow}$' axis x1y2
        
    % \end{gnuplot}
    \captionsetup{font=footnotesize}
    \vspace{-10pt}
    \caption{Effectiveness of $\gamma$ on LM}
    \vspace{-20pt}
    \label{fig:gamma-lm}

%% file: tables/bart_perf_sncr.tex
\vspace{-23pt}
\centering
\captionsetup{font=footnotesize}
\caption{Effectiveness of $\sncr$}
\label{table:ablation:sncr}
\vspace{-8pt}
\resizebox{0.46\columnwidth}{!}{%
\bgroup
\def\arraystretch{1}%
\begin{tabular}{
      @{\hspace{5pt}}l@{\hspace{15pt}}
      @{\hspace{0pt}}r@{\hspace{8pt}}
      @{\hspace{0pt}}r@{\hspace{8pt}}
      @{\hspace{0pt}}r@{\hspace{8pt}}
      @{\hspace{0pt}}r@{\hspace{8pt}}
      @{\hspace{0pt}}r@{\hspace{8pt}}
      @{\hspace{0pt}}r@{\hspace{5pt}}
}

\toprule

$\sncr$  &  F1    & ROUGE1 & BLEU & PPL$^{\downarrow}$ &  PG &  KG  \\

\midrule

0.25  &  0.359 & 0.427 & 15.744 &    8.02 & 71.75  & 95.69    \\
0.35  &  \bf{0.360} & \bf{0.429} & 15.768 &    \bf{7.97} & 65.22  & \bf{96.35}    \\
0.50  &  0.359 & 0.428 & \bf{15.854} &    8.12 & 67.83  & 96.15    \\
0.75  &  0.355 & 0.425 & 15.480 &    8.02 & \bf{72.23}  & 96.33    \\
1.00  &  0.348 & 0.410 & 14.642 &    9.19 & 51.82  & 90.42    \\

\bottomrule
\end{tabular}
\egroup
}
\vspace{-20pt}

%% file: sections/conclusion.tex
\section{Conclusions}

Conversational agents built on DNNs and LLMs are playing increasingly critical roles in numerous real-world applications. For increased adoption in the future, it is crucial that these agents consistently provide truthful responses and tailor them to users' personal preferences. By leveraging auxiliary data like external knowledge and user personas, these agents are becoming increasingly powerful and versatile. However, effectively and efficiently retaining relevant information while filtering out irrelevant data remains a challenging problem.

In this paper, we presented a novel method, \ourmethod, that is able to (i) encode multiple auxiliary data sources and preserve source-specific signals in specialized streams, (ii) accurately identify relevant persona and knowledge entries in two post-fusion grounding networks via a novel text similarity measure ($\ssncolbertsim$) at the word-level with normalization, symmetry and sparsity, and (iii) generate high-quality responses that respect provided facts and that are personalized to a user.
Our experiments demonstrated that \ourmethod significantly outperforms the SOTA methods, \pkfocus and \pkncli, in terms of both PG/KG accuracies and language generation quality. We also conducted a comprehensive study on specific choices within \ourmethod, including the weights of various components in the model, the choice of language model, and sparse sampling strategies. This paper demonstrated the importance of leveraging normalization, sparsity and symmetry among multiple auxiliary sources to effectively extract useful information and provide supplementary view for the conversational context.
% Furthermore, we identified potential avenues for future improvements to our methods, such as exploring more effective techniques for sampling informative tokens during the generation process.

\newcontent{Our method, \ourmethod, still faces challenges. A major challenge observed in both the baselines and \ourmethod is hallucination, where generated responses contain inaccuracies and misinformation despite appearing genuine. Another issue with \ourmethod is that its efficiency is limited by the online calculation of TF-IDF weights, particularly when novel tokens are present in the query in real-world applications. Furthermore, \ourmethod was only evaluated on the FoCus dataset due to the lack of appropriate publicly available evaluation datasets. We will focus future research efforts on addressing these challenges. 
}